\documentclass[journal]{IEEEtran}
\makeatletter\if@twocolumn\PassOptionsToPackage{switch}{lineno}\else\fi\makeatother

\usepackage{graphicx}
\usepackage[T1]{fontenc}
\usepackage[numbers,sort&compress]{natbib}
\usepackage{algorithm,algpseudocode}
\usepackage{pgfplots}
\pgfplotsset{width=7cm}
\usepackage{subcaption}
\usepackage{longtable}
\usepackage{algpseudocode}
\usepackage{varwidth}
\ifCLASSINFOpdf
\else
\fi
\usepackage{amsmath}
\usepackage[cmintegrals]{newtxmath}
\usepackage{url,multirow,morefloats,floatflt,cancel,tfrupee}
\makeatletter

\AtBeginDocument{\@ifpackageloaded{textcomp}{}{\usepackage{textcomp}}}
\makeatother
\usepackage{colortbl}
\usepackage{xcolor}
\usepackage{pifont}
\usepackage[nointegrals]{wasysym}
\makeatletter

\def\mcWidth#1{\csname TY@F#1\endcsname+\tabcolsep}

\def\cAlignHack{\rightskip\@flushglue\leftskip\@flushglue\parindent\z@\parfillskip\z@skip}
\def\rAlignHack{\rightskip\z@skip\leftskip\@flushglue \parindent\z@\parfillskip\z@skip}

\usepackage{ifxetex}
\ifxetex\else\if@twocolumn\usepackage{dblfloatfix}\fi\fi

\AtBeginDocument{
\expandafter\ifx\csname eqalign\endcsname\relax
\def\eqalign#1{\null\vcenter{\def\\{\cr}\openup\jot\m@th
  \ialign{\strut$\displaystyle{##}$\hfil&$\displaystyle{{}##}$\hfil
      \crcr#1\crcr}}\,}
\fi
}

\AtBeginDocument{%
  \@ifpackageloaded{endfloat}%
   {\renewcommand\efloat@iwrite[1]{\immediate\expandafter\protected@write\csname efloat@post#1\endcsname{}}}{}%
}%

\def\BreakURLText#1{\@tfor\brk@tempa:=#1\do{\brk@tempa\hskip0pt}}
\let\lt=<
\let\gt=>
\def\processVert{\ifmmode|\else\textbar\fi}

\@ifundefined{subparagraph}{
\def\subparagraph{\@startsection{paragraph}{5}{2\parindent}{0ex plus 0.1ex minus 0.1ex}%
{0ex}{\normalfont\small\itshape}}%
}{}

\newcommand\role[1]{\unskip}
\newcommand\aucollab[1]{\unskip}
  
\@ifundefined{tsGraphicsScaleX}{\gdef\tsGraphicsScaleX{1}}{}
\@ifundefined{tsGraphicsScaleY}{\gdef\tsGraphicsScaleY{.9}}{}
\def\checkGraphicsWidth{\ifdim\Gin@nat@width>\linewidth
	\tsGraphicsScaleX\linewidth\else\Gin@nat@width\fi}

\def\checkGraphicsHeight{\ifdim\Gin@nat@height>.9\textheight
	\tsGraphicsScaleY\textheight\else\Gin@nat@height\fi}

\def\fixFloatSize#1{}%\@ifundefined{processdelayedfloats}{\setbox0=\hbox{\includegraphics{#1}}\ifnum\wd0<\columnwidth\relax\renewenvironment{figure*}{\begin{figure}}{\end{figure}}\fi}{}}
\let\ts@includegraphics\includegraphics

\def\inlinegraphic[#1]#2{{\edef\@tempa{#1}\edef\baseline@shift{\ifx\@tempa\@empty0\else#1\fi}\edef\tempZ{\the\numexpr(\numexpr(\baseline@shift*\f@size/100))}\protect\raisebox{\tempZ pt}{\ts@includegraphics{#2}}}}

\AtBeginDocument{\def\includegraphics{\@ifnextchar[{\ts@includegraphics}{\ts@includegraphics[width=\checkGraphicsWidth,height=\checkGraphicsHeight,keepaspectratio]}}}

\def\URL#1#2{\@ifundefined{href}{#2}{\href{#1}{#2}}}

\def\UrlOrds{\do\*\do\-\do\~\do\'\do\"\do\-}%
\g@addto@macro{\UrlBreaks}{\UrlOrds}

\@ifundefined{quoteAttrib}
	{}
	{}

\@ifundefined{titlequoteAttrib}
	{}{}

\newenvironment{title-quote}
	{\list{}{\fontsize{10pt}{12pt}\selectfont\leftmargin.5in\itshape\rightmargin\leftmargin}%
  \item\relax}
  {\endlist}

\makeatother

\usepackage{tabulary}
\makeatletter
\AtBeginDocument{\@ifpackageloaded{longtable}{%
\def\LT@makecaption#1#2#3{%
  \LT@mcol\LT@cols c{\hbox to\z@{\hss\parbox[t]\LTcapwidth{%
    \sbox\@tempboxa{#1{#2: } #3}%
    \ifdim\wd\@tempboxa>\hsize
      #1{#2: }\textsc{#3}%
    \else
      \hbox to\hsize{\hfil\box\@tempboxa\hfil}%
    \fi
    \endgraf\vskip\baselineskip}%
  \hss}}}
}{}}
\makeatother

   \makeatletter
  \def\fig@textbf{\textbf}
   \AtBeginDocument{\renewcommand\floatc@plain[2]{\setbox\@tempboxa\hbox{{\footnotesize#1.}\footnotesize\hskip.5em#2}%
    \ifdim\wd\@tempboxa>\hsize {\fig@textbf{\footnotesize#1.}}\footnotesize\hskip.5em#2\par
        \else\hbox to\hsize{\hfil\box\@tempboxa\hfil}\fi}}
    \makeatother
  
\usepackage{float}

\usepackage{listings,xcolor}

\begin{document}


        \title{Ask less {\textemdash} Scale Market Research without Annoying Your Customers}
      
\author{Venkatesh~Umaashankar and 
        Girish~Shanmugam S\thanks{Venkatesh~Umaashankar is with Ericsson Research,
        Chennai,
        India, e-mail: venkatesh.u@ericsson.com (Corresponding author).}\thanks{Girish~Shanmugam S is a Machine Learning Consultant,
        E3, Jains Green Acres,
        Chennai,
        India, e-mail: s.girishshanmugam@gmail.com}}

\maketitle 

\begin{abstract}
Market research is generally performed by surveying a representative sample of customers with questions that includes contexts such as psycho-graphics, demographics, attitude and product preferences. Survey responses are used to segment the customers into various groups that are useful for targeted marketing and communication. Reducing the number of questions asked to the customer has utility for businesses to scale the market research to a large number of customers. In this work, we model this task using Bayesian networks. We demonstrate the effectiveness of our approach using an example market segmentation of broadband customers.
\end{abstract}
    

% \fi
\IEEEpeerreviewmaketitle

\section{Introduction}
A key technique for developing successful business strategies in business to customer (B2C) companies is to develop a good understanding of the market and the customer behavior. Market research and segmentation play an important role in framing business and marketing strategies, which help organizations to improve the efficiency of their marketing and conversion. Market segmentation could be defined as the process of breaking down the market for a particular product or service into segments of customers which differ in terms of their response to marketing strategies\unskip~\cite{check1}.

Market segmentation comprises of 2 major steps. (1) Consumer Survey --- A survey questionnaire considering various dimensions such as psycho-graphics, demographics, attitude, product usage, preferences is meticulously designed. Psycho-graphic questions are useful in understanding the preferences and behavior of customers\unskip~\cite{297803:6606163}. The carefully planned survey is then rolled out to a representative sample of customers. (2) Segment Generation --- The survey responses are analyzed to create a segmentation model. The segmentation model could be rule, algorithm or factor analysis based. This model can be abstractly defined as a function that could be used to assign a segment to every surveyed customer. The process of market segmentation is discussed in detail in\unskip~\cite{connect-ds-marketsegmentation}.
\begin{equation} \label{eq:1}
    SegmentationModel(responses) = segment
\tag{1}\end{equation}
The market segments are summarized by profiles and are given descriptive names. Consider the example of market segmentation of movie consumers\unskip~\cite{297803:6606161} shown in Figure~\ref{figure-93c8c941b321ff3cd1eed19b8e7e650c}. This market research was done for a studio to understand the level of piracy among movie consumers. They report four clear market segments classified based on consumption level and tendency to consume pirated material.

\bgroup
\fixFloatSize{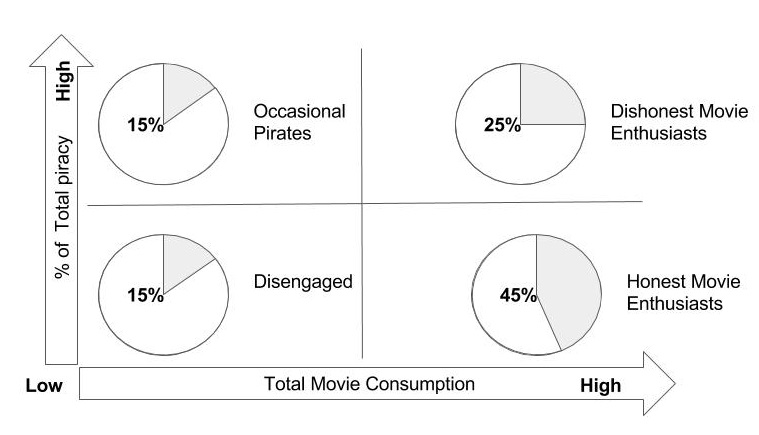}
\begin{figure}[!htbp]
\centering \makeatletter\IfFileExists{images/72c18482-8100-47e7-80dc-7d48ed65774a-umovie_consumers_segmentation.jpg}{\includegraphics{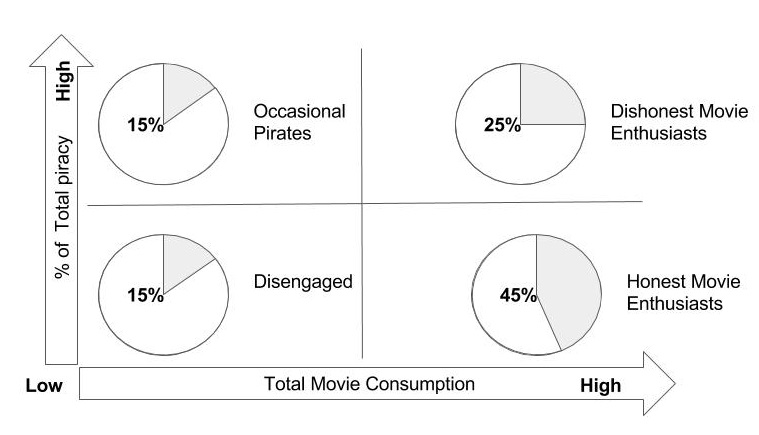}}{}
\makeatother 
\caption{{Movie Consumers Market Segmentation}}
\label{figure-93c8c941b321ff3cd1eed19b8e7e650c}
\end{figure}
\egroup

\begin{table*}[!htbp]
\caption{{ Segment Descriptions} }
\label{table-wrap-1fe8a3603d86cf7bb46f37ef5b523dcb}
\def\arraystretch{1}
\ignorespaces 
\centering 
\begin{tabulary}{\linewidth}{LLLL}
\hline 
 \textbf{Occasional Pirates} & \textbf{Disengaged} & \textbf{Dishonest Enthusiasts} & \textbf{Honest Enthusiasts}\\
\hline 
Higher conversion potential &
  Moderate conversion Potential &
  Low conversion &
  High potential\\ 
Pirating family / action \& Low risk to pirate &
  Low risk to pirate &
  Action preferred genre &
  Across all genres\\
females 25-34 with young family &
  45+ family &
  younger male students &
  35-44 with family\\
\hline 
\end{tabulary}\par 
\end{table*}

The heterogeneity among the segments is emphasized in the descriptions are shown in Table~\ref{table-wrap-1fe8a3603d86cf7bb46f37ef5b523dcb}. Segment descriptions help to build an intuition about the nature and behavior of each segment. Market segmentation for a product or service is usually executed by expert market research companies\footnote{Ipsos and TNS are well-known market research experts in the industry}. The key outcomes of the market research are segmentation model, target segments, presentations and workshops to spread the awareness within the organization. Market segmentation has been battle-tested in many consumer-facing business and it clearly helps to build the intuition about the big picture. Still, it is an open challenge to scale market research to millions of customers.  It is not practical to ask a long list of questions to each and every customer, that would not only be time-consuming but also be annoying the customers.

Factor analysis is a well-known method for estimating the latent traits from question-level survey data and to reduce the number of questions \unskip~\cite{297803:6606170}. However, it has also been the subject of no small amount of criticism among market researchers \unskip~\cite{297803:6606172}. The major problem with factor analysis is that we loose the diversity in the collected information, and we will have only minimal information. A factor analysis carried out on one-half of the data might give different results from those obtained from the other half, thus making the reliability of results questionable. Yet another limitation is that it is unable to give a unique solution or result. An exercise in factor analysis involving a large number of variables say 50, is much bothersome, costly and time-consuming \unskip~\cite{297803:6606173}. Due to these limitations, we avoided factor analysis and decided to go for a much simpler alternative.

A Bayesian framework that systematically addresses the challenges faced when the future value of customers is estimated based on survey data has been proposed in \unskip~\cite{297803:6606174}. A method for building effective Bayesian network (BN) models for medical decision support from complex, unstructured and incomplete patient questionnaires and interviews was developed in \unskip~\cite{297803:6606180}. It extends to challenging the decision scientists to reason about building models based on what information is really required for inference.

The closest to our work is \unskip~\cite{297803:6606177} where Bayesian network modeling has been used instead of applying factor analysis technique to determine key factors from a survey questionnaire, to find the most accurate representation of the complex system and identify key variables for understanding the subsequent effects of blast exposure based on an online survey. To the best of our knowledge, there has not been any other work exploring the use of the Bayesian network for scaling market research or to reduce the number of questions in a market research survey.

In this work, we propose a novel way to use Bayesian Networks to reduce the number of questions that a customer needs to be asked. In addition to that, we demonstrate the effectiveness of our approach by evaluating the segment assigned by the Bayesian Network model when fewer questions are asked in the survey. Finally, we summarize the advantages of our approach and discuss our conclusions.

\section{Proposed Approach}
Inspired by the success of using Bayesian networks to understand and analyze survey data \unskip~\cite{297803:6606171}, we propose a Bayesian Network based approach for reducing the number of questions in a market research survey. The outline of our approach is shown in Figure~\ref{figure-98ad891a0421ff226922d93ad4c29178}. Our approach consists of two phases: (1) Preparatory Phase and (2) Scaling Phase.

\bgroup
\fixFloatSize{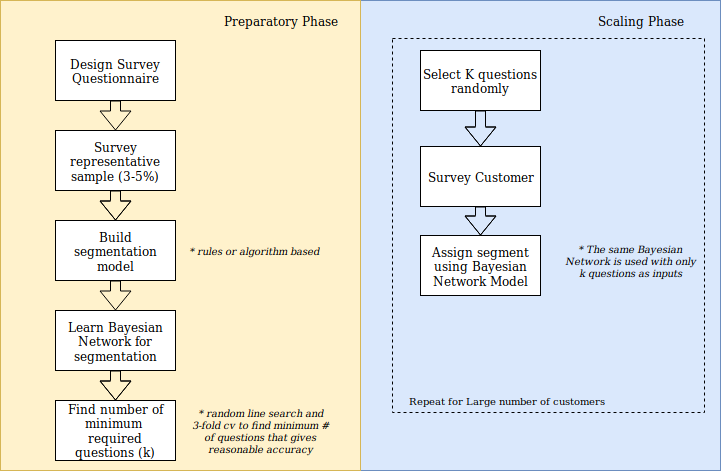}
\begin{figure*}[!htbp]
\centering \makeatletter\IfFileExists{images/f6a0c1f4-94a0-434e-afa1-6385fce5df5d-uaskless.png}{\includegraphics[scale=0.75,keepaspectratio]{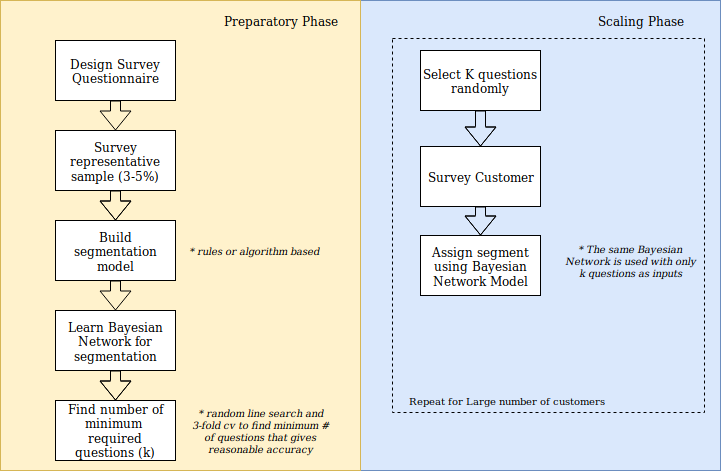}}{}
\makeatother 
\caption{{Ask less - Approach outline}}
\label{figure-98ad891a0421ff226922d93ad4c29178}
\end{figure*}
\egroup

\begin{algorithm}
\scriptsize
\caption{Find optimal number of minimum questions}\label{alg:findk}
\begin{algorithmic}
\Function{Find\_k}{\null}
 \State $K\gets [5,10,15,..]$\Comment{line search to find optimal value for $k$}
 \State $bestFscore \gets 0.0$
 \State $bestK \gets None$
 \For{$k \in K$}
  \State $segments \gets dict()$
   \For{each customer in $testSet$}
     \State $questions\gets randomQuestions(k)$
     \State $responses~\gets~surveyResponses($
     \Statex[9] $questions,customers)$
     \State $segments[customer]~\gets~bayesianNetworkModel($
     \Statex[11] $responses)$
       \EndFor
   \State $metric \gets fscore(segments)$
   \If {$metric > bestFscore$}
      \State $bestFscore \gets metric$
      \State $bestK \gets k$
    \EndIf
\EndFor
\State \Return $bestK$
\EndFunction
\end{algorithmic}
\end{algorithm}

\subsection{Preparatory phase}
\label{ssec:pphase}In the Preparatory phase, the survey questionnaire is designed and the survey is rolled out to a representative sample of customers, which is typically 2-5\% of the total customer base. Customers selected for these phases are usually sampled in stratified fashion, across various regions of value that they add to the business. The survey responses are analyzed and a segmentation model is built to divide the customers into different segments. A segmentation model could be defined as a function that takes survey responses as input and provides customer segment as output as shown in the equation \ref{eq:1}. Note that the questions have to be carefully designed, keeping in mind what type of segments the business would benefit from. Also, note that it might be the case that all the survey responses are similar and it might not be possible to differentiate customers based on the responses. In such cases, one has to iterate again to identify the suitable questions and customers.

The next key step is to learn a Bayesian Network model that \textbf{\textit{approximates the segmentation model}}. All the questions in the survey questionnaire and the segment are represented as nodes in this Bayesian network. Learning a Bayesian Network model involves two steps: (1) Structure learning --- A Bayesian network is represented by a directed acyclic graph (DAG). The DAG structure could be learnt with either score-based approach or constraint-based approach. The score-based approach first defines a criterion to evaluate how well the Bayesian network fits the data e.g BIC Score, then searches over the space of DAGs for a structure with maximum score \unskip~\cite{297803:6606175}\unskip~\cite{297803:6606176}. The constraint-based case uses the independence test to identify a set of edge constraints for the graph and then finds the best DAG that satisfies the constraints \unskip~\cite{297803:6606178}\unskip~\cite{297803:6606179}. (2) Parameter Learning --- This involves learning the parameters that are required to estimate the conditional probability tables of each node in the Bayesian network. These parameters are typically learned through Expectation maximization, Maximum likelihood, and gradient-based approaches. We use 70\% of the survey data to learn the Bayesian Network model.

 A key advantage of a Bayesian Network model is its ability to handle partial information at the time of inference i.e the same Bayesian Network model could be used for segment assignment even when fewer questions are asked. The main novelty in our approach is to exploit this property of Bayesian Networks to reduce the number of questions in the survey. We find an optimal hyper parameter \textit{k}, which is the number of random questions that could be asked to the customer whose responses when fed to the Bayesian Network model will guarantee an average f-score above a configured threshold for example  0.70. In simple terms, how many fewer questions I could ask without compromising too much on the Bayesian Network segmentation model performance. The algorithm that we used to identify the optimal number of minimal questions is shown in Algorithm \ref{alg:findk}.

\subsection{Scaling Phase}
\label{ssec:scaling}Once the optimal value for \textit{k} has been identified as explained in the previous section, the scaling phase becomes very simple. A customer gets asked only \textit{k} random questions, instead of going through the whole questionnaire. The responses to these \textit{k} questions are passed through the Bayesian Network model and the segment assignment is done. This approach also provides an opportunity for incrementally updating the segment assignment as new information becomes available. For example, the customer can also be questioned in multiple parts and the segment assigned to the customer can be updated based on his additional responses.

\bgroup
\fixFloatSize{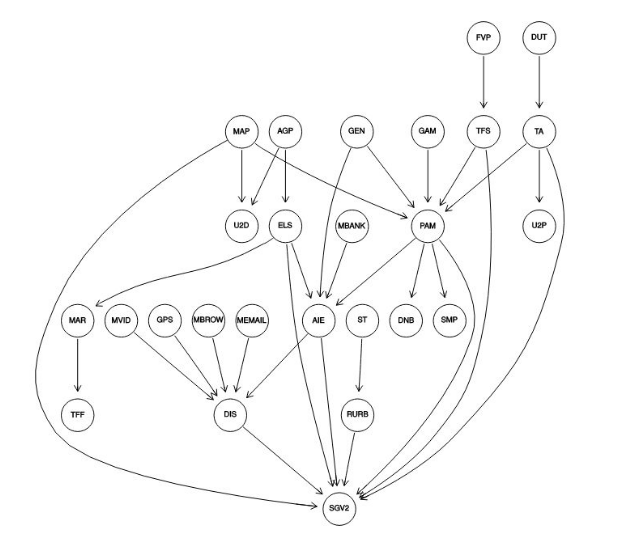}
\begin{figure}[!htbp]
\centering \makeatletter\IfFileExists{images/segmentation_network.png}{\includegraphics[height=3.3in, width=3.5in]{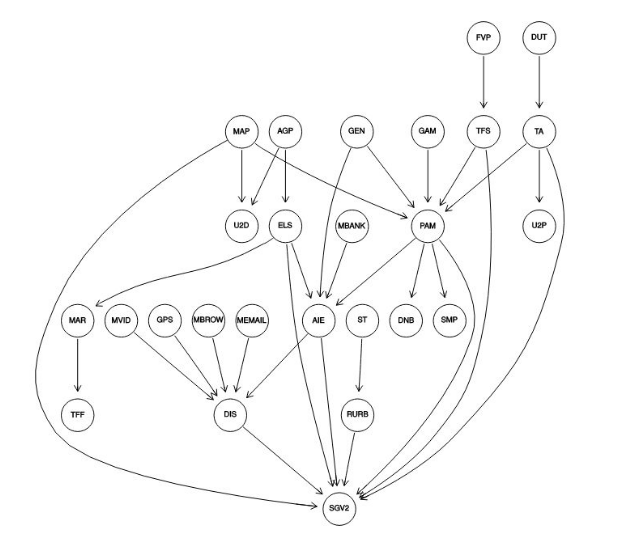}}{}
\makeatother 
\caption{{Ask Less - Bayesian Network Model}}
\label{fig:segmentation_network}
\end{figure}
\egroup

\section{Results}
We implemented our proposed approach to scale the market research that was performed for an Internet Service Provider (ISP) business. A total of 100,000 customers participated in the survey. The survey participants were sampled from the total customers based on their plan and lifetime value in a stratified manner. Most of the survey questions are scale based (1 to 5), a response of 1 means the participant strongly disagrees with the statement in the question whereas a response of 5 means that the participant strongly agrees with the statement. A complete list of survey questions is shown in the Table~\ref{table-wrap-85100297486465213d10441f1141e979}. The survey responses were analyzed and a combination of rule and algorithm based segmentation model was built and 4 customer segments (S1, S2, S3, and S4) were identified.

In the Preparatory phase \ref{ssec:pphase} described in our approach, we learnt the structure of the Bayesian network using Hill-Climbing (hc) greedy search on the space of directed graph and Akaike Information Criterion (AIC) as the scoring criteria. We used the Maximum-Likelihood estimates for fitting the parameters of the Bayesian Network. For both structure learning and parameter fitting, we used the implementation available in the bnlearn R package \unskip~\cite{297803:6606182}.  Figure. \ref{fig:segmentation_network} shows the structure of our Bayesian Network model. We used 70\% (70,000) of the survey responses to learn the Bayesian Network Model. Note that the nodes in the model are responses to survey questions and the corresponding segment assignment for the customer (SGV2). The learned network structure was validated with domain experts, and we list few interesting observations: (1) A person's perception about mobile (PAM) influences if he wants to access internet everywhere (AIE). (2) The final segment assigned to the customer is based on the fact if that customer uses diverse internet services (DIS). (3) Gender of the customer (GEN) could influence the customer's perception about mobile (PAM) and his urge to access internet everywhere (AIE). (4) The customer's value for features in a product (FVP) decides if he wants to use the product to showoff (TFS).

We used the line search algorithm shown in Algorithm.~\ref{alg:findk} to identify the optimal hyper parameter \textit{k}, which is the number of random questions that could be asked to the customer that will guarantee an average f-score above 0.70. We used 30\% of the survey responses (30,000) for this purpose. We have a total of 22 questions in the survey. We ran the \textit{Find\_k} algorithm with values for k as [5,10,20]. The segment classification performance metrics of Bayesian Network model for each value of \textit{k} is shown in Table~\ref{table-wrap-180296f8afe618e2dddd5165b121df31} , Table~\ref{table-wrap-30e8d16fd1697c97253e5399207c4a53} and Table~\ref{table-wrap-cb0701f61736b6c1f061ad117351d5f5}.  We use the cpquery function of bnlearn to supply the partial evidence i.e responses for randomly selected questions to run a conditional probability query and predict the segment assignment. We found the optimal value for \textit{k} is 10 in this case. This means that by using our approach, we could reduce the number of questions by 50\%. Figure \ref{fig:f_score_comparison_graph} shows the comparison of scores for various values of \textit{k}.

In the Scaling phase \ref{ssec:scaling}, we integrate our Bayesian Network model with the survey tool which randomly selects \textit{k} (10) questions and collects the responses for them from the customers. These responses are passed as evidence to the Bayesian Network model and segment are assigned.

\begin{table}[!htbp]
\caption{{ Accuracy Metrics for k=20} }
\label{table-wrap-180296f8afe618e2dddd5165b121df31}
\def\arraystretch{1}
\ignorespaces 
\centering 
\begin{tabulary}{\linewidth}{LLLL}
\hline 
Segment & Precision & Recall & F-Score\\
\hline 
S1 &
  0.87 &
  0.90 &
  0.89\\
S2 &
  0.68 &
  0.79 &
  0.73\\
S3 &
  0.85 &
  0.86 &
  0.86\\
S4 &
  0.96 &
  0.81 &
  0.88\\
Average &
  0.86 &
  0.85 &
  0.85\\
\hline 
\end{tabulary}\par 
\end{table}

\begin{table}[!htbp]
\caption{{ Accuracy Metrics for k=10} }
\label{table-wrap-30e8d16fd1697c97253e5399207c4a53}
\def\arraystretch{1}
\ignorespaces 
\centering 
\begin{tabulary}{\linewidth}{LLLL}
\hline 
Segment & Precision & Recall & F-Score\\
\hline 
S1 &
  0.77 &
  0.87 &
  0.81\\
S2 &
  0.52 &
  0.64 &
  0.57\\
S3 &
  0.76 &
  0.65 &
  0.70\\
S4 &
  0.91 &
  0.76 &
  0.83\\
Average &
  0.75 &
  0.74 &
  0.74\\
\hline 
\end{tabulary}\par 
\end{table}

\begin{table}[!htbp]
\caption{{ Accuracy Metrics for k=5} }
\label{table-wrap-cb0701f61736b6c1f061ad117351d5f5}
\def\arraystretch{1}
\ignorespaces 
\centering 
\begin{tabulary}{\linewidth}{LLLL}
\hline 
Segment & Precision & Recall & F-Score\\
\hline 
S1 &
  0.67 &
  0.76 &
  0.71\\
S2 &
  0.38 &
  0.52 &
  0.44\\
S3 &
  0.65 &
  0.53 &
  0.59\\
S4 &
  0.79 &
  0.63 &
  0.70\\
Average &
  0.65 &
  0.62 &
  0.63\\
\hline 
\end{tabulary}\par 
\end{table}

\begin{center}
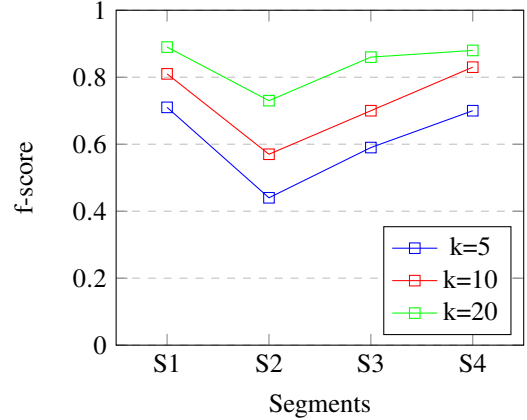

\begin{tikzpicture}
\begin{axis}[
    xlabel={Segments},
    ylabel={f-score},
    xmin=0, xmax=40,
    ymin=0, ymax=1,
    xtick={5,15,25,35},
    ytick={0,0.20,0.40,0.60,0.80,1},
    xticklabels={S1,S2,S3,S4},
    legend pos=north west,
    ymajorgrids=true,
    grid style=dashed,
    legend pos=south east
]

\addplot[
    color=blue,
    mark=square,
    ]
    coordinates {
    (5,0.71)(15,0.44)(25,0.59)(35,0.70)
    };
    \addlegendentry{k=5}
\addplot[
    color=red,
    mark=square,
    ]
    coordinates {
    (5,0.81)(15,0.57)(25,0.70)(35,0.83)
    };
    \addlegendentry{k=10}
 \addplot[
    color=green,
    mark=square,
    ]
    coordinates {
    (5,0.89)(15,0.73)(25,0.86)(35,0.88)
    };
    \addlegendentry{k=20}
    
\end{axis}
\end{tikzpicture}
\captionof{figure}{Comparison of f-scores for different 'k' values across segments}
\label{fig:f_score_comparison_graph}
\end{center}

\begin{table*}[!htbp]
\caption{{ Survey Questionnaire} }
\label{table-wrap-85100297486465213d10441f1141e979}
\def\arraystretch{1}
\ignorespaces 
\centering 
\begin{tabulary}{\linewidth}{|p{0,3cm}|p{1,4cm}|p{3cm}|p{9cm}|}
\hline 
Id & ABBR & Expansion & Actual Question\\
\hline 
1 &
  AGP &
  Age group &
  \\\cline{1-1}\cline{2-2}\cline{3-3}\cline{4-4}
2 &
  MAR &
  Marital Status &
  \\\cline{1-1}\cline{2-2}\cline{3-3}\cline{4-4}
3 &
  PAM &
  Perception About Mobile &
  I find new technology exciting and want to have a mobile phone with the latest services and features.\\\cline{1-1}\cline{2-2}\cline{3-3}\cline{4-4}
4 &
  AIE &
  Access Internet Everywhere &
  It's important for me to be able to access the Internet wherever I am\\\cline{1-1}\cline{2-2}\cline{3-3}\cline{4-4}
5 &
  MAP &
  Most Advanced Products &
  I'm constantly looking for the most technologically advanced products available\\\cline{1-1}\cline{2-2}\cline{3-3}\cline{4-4}
6 &
  DUT &
  Difficulty in Using Technology &
  For me to use a new technology product, somebody has to show me how to use it\\\cline{1-1}\cline{2-2}\cline{3-3}\cline{4-4}
7 &
  TA &
  Technology Avert &
  I feel that I am able to manage without many of the technology products that other people find essential\\\cline{1-1}\cline{2-2}\cline{3-3}\cline{4-4}
8 &
  FVP &
  Features Vs Price &
  The features are more important than the price\\\cline{1-1}\cline{2-2}\cline{3-3}\cline{4-4}
9 &
  U2D &
  Up To Date &
  It is important to be up\-to\-date on major news\\\cline{1-1}\cline{2-2}\cline{3-3}\cline{4-4}
10 &
  TFS &
  Technology For Showoff &
  Carrying the latest technology products makes a good impression\\\cline{1-1}\cline{2-2}\cline{3-3}\cline{4-4}
11 &
  U2P &
  Unwilling To Pay &
  Even when I can afford them, I'm not willing to pay much for new technology products or services\\\cline{1-1}\cline{2-2}\cline{3-3}\cline{4-4}
12 &
  DNB &
  Dont Need Mobile &
  I do not need a mobile phone\\\cline{1-1}\cline{2-2}\cline{3-3}\cline{4-4}
13 &
  MBROW &
  Mobile Browsing &
  Mobile Browsing of the Internet\\\cline{1-1}\cline{2-2}\cline{3-3}\cline{4-4}
14 &
  MEMAIL &
  Mobile Email &
  Send and Receive Email via the mobile phone\\\cline{1-1}\cline{2-2}\cline{3-3}\cline{4-4}
15 &
  MBANK &
  Mobile Banking &
  Mobile Banking via the mobile phone.\\\cline{1-1}\cline{2-2}\cline{3-3}\cline{4-4}
16 &
  MVID &
  Mobile Video &
  Watching videos on your mobile phone\\\cline{1-1}\cline{2-2}\cline{3-3}\cline{4-4}
17 &
  GPS &
  Global Position Tracking &
  Mapping, navigation or positioning service (like gps) via the mobile phone\\\cline{1-1}\cline{2-2}\cline{3-3}\cline{4-4}
18 &
  GAM &
  Gaming &
  Playing video games is one of my favourite activities\\\cline{1-1}\cline{2-2}\cline{3-3}\cline{4-4}
19 &
  SMP &
  Small Payments &
  Small Payment service via the mobile phone\\\cline{1-1}\cline{2-2}\cline{3-3}\cline{4-4}
20 &
  TFF &
  Time For Family &
  I spend a lot of time with my family\\\cline{1-1}\cline{2-2}\cline{3-3}\cline{4-4}
21 &
  RURB &
  Rural or Urban &
  \\\cline{1-1}\cline{2-2}\cline{3-3}\cline{4-4}
22 &
  ELS &
  Life Stage &
  \\\cline{1-1}\cline{2-2}\cline{3-3}\cline{4-4}
23 &
  DIS &
  Diverse Internet Services &
  Derived Attribute\\\cline{1-1}\cline{2-2}\cline{3-3}\cline{4-4}
24 &
  SGV2 &
  Segment Labels &
  \\\cline{1-1}\cline{2-2}\cline{3-3}\cline{4-4}
\hline 
\end{tabulary}\par 
\end{table*}

\section{Conclusion}
In this paper, we propose a simpler way to reduce the number of questions in a Market Research survey using Bayesian networks. We evaluated the effectiveness of our approach in a real-world setting, and we observe that our approach can help to reduce up to 50\% of the questions with a minor dip in classification performance. Our work shows that Bayesian networks can serve as a simpler alternative to factor analysis to reduce the number of questions in a survey, without compromising the ability to collect information about various topics.

\section*{Acknowledgments}We thank Prasad Garigipati, Henrik Palson, Andreas Timglas and Roy Ollila for their help and support. Both the authors got introduced to the area of Market Research during their tenure at Xoanon Analytics. The value in asking fewer questions in a Market Research Survey was recognized by the authors based on their practical experience.



%\bibliographystyle{IEEEtran}

\bibliographystyle{IEEEtran}
\bibliography{references}

\begin{thebibliography}{10}
\providecommand{\url}[1]{#1}
\csname url@samestyle\endcsname
\providecommand{\newblock}{\relax}
\providecommand{\bibinfo}[2]{#2}
\providecommand{\BIBentrySTDinterwordspacing}{\spaceskip=0pt\relax}
\providecommand{\BIBentryALTinterwordstretchfactor}{4}
\providecommand{\BIBentryALTinterwordspacing}{\spaceskip=\fontdimen2\font plus
\BIBentryALTinterwordstretchfactor\fontdimen3\font minus
  \fontdimen4\font\relax}
\providecommand{\BIBforeignlanguage}[2]{{%
\expandafter\ifx\csname l@#1\endcsname\relax
\typeout{** WARNING: IEEEtran.bst: No hyphenation pattern has been}%
\typeout{** loaded for the language `#1'. Using the pattern for}%
\typeout{** the default language instead.}%
\else
\language=\csname l@#1\endcsname
\fi
#2}}
\providecommand{\BIBdecl}{\relax}
\BIBdecl

\bibitem{check1}
Y.~Wind and S.~P. Douglas, ``{International market segmentation},''
  \emph{{European Journal of Marketing}}, vol.~6, no.~1, pp. 17--25, 1972.

\bibitem{297803:6606163}
N.~M. Bradburn, S.~Sudman, and B.~Wansink, \emph{{Asking questions: the
  definitive guide to questionnaire design-for market research, political
  polls, and social and health questionnaires}}.\hskip 1em plus 0.5em minus
  0.4em\relax John Wiley \& Sons, 2004.

\bibitem{connect-ds-marketsegmentation}
L.~Cremonezi, ``High definition customers - a powerful segmentation,'' Ipsos
  MORI, White Paper, 2016.

\bibitem{297803:6606161}
Z.~Andrew and D.~Peter, ``{A Guide to getting the best out of your Segmentation
  Analyses},'' 2011.

\bibitem{297803:6606170}
R.~D.~F. Jr, W.~W. Kulzy, and J.~A. Appleget, ``{From data to information:
  Using factor analysis with survey data},'' \emph{{Phalanx}}, vol.~45, no.~4,
  pp. 30--34, 2012.

\bibitem{297803:6606172}
A.~S.~C. Ehrenberg, G.~J. Goodhardt, and S.~I. Marketing, \emph{{Factor
  analysis: limitations and alternatives}}.\hskip 1em plus 0.5em minus
  0.4em\relax Marketing Science Institute.

\bibitem{297803:6606173}
G.~C. Beri, \emph{{Marketing research}}.\hskip 1em plus 0.5em minus 0.4em\relax
  Tata McGraw-Hill Education, 2007.

\bibitem{297803:6606174}
J.~Karvanen, A.~Rantanen, and L.~Luoma, ``{Survey data and Bayesian analysis: a
  cost-efficient way to estimate customer equity},'' \emph{{Quantitative
  Marketing and Economics}}, vol.~12, no.~3, pp. 305--329, 2014.

\bibitem{297803:6606180}
A.~C. Constantinou, N.~Fenton, W.~Marsh, and L.~Radlinski, ``{From complex
  questionnaire and interviewing data to intelligent Bayesian network models
  for medical decision support},'' \emph{{Artificial intelligence in
  medicine}}, vol.~67, pp. 75--93, 2016.

\bibitem{297803:6606177}
P.~A. Toyinbo, R.~D. Vanderploeg, H.~G. Belanger, A.~M. Spehar, W.~A. Lapcevic,
  and S.~G. Scott, ``{A systems science approach to understanding polytrauma
  and blast-related injury: Bayesian network model of data from a survey of the
  Florida National Guard},'' \emph{{American journal of epidemiology}}, vol.
  185, no.~2, pp. 135--146, 2017.

\bibitem{297803:6606171}
S.~Salini and R.~S. Kenett, ``{Bayesian networks of customer satisfaction
  survey data},'' \emph{{Journal of Applied Statistics}}, vol.~36, no.~11, pp.
  1177--1189, 2009.

\bibitem{297803:6606175}
N.~Friedman, K.~Murphy, and S.~Russell, ``{Learning the structure of dynamic
  probabilistic networks},'' in \emph{{Proceedings of the Fourteenth conference
  on Uncertainty in artificial intelligence}}, 1998, pp. 139--147.

\bibitem{297803:6606176}
D.~Heckerman, D.~Geiger, and D.~M. Chickering, ``{Learning Bayesian networks:
  The combination of knowledge and statistical data},'' \emph{{Machine
  learning}}, vol.~20, no.~3, pp. 197--243, 1995.

\bibitem{297803:6606178}
H.~Steck, ``{Constraint-based structural learning in Bayesian networks using
  finite data sets},'' 2001.

\bibitem{297803:6606179}
C.~P. de~Campos and Q.~Ji, ``{Efficient structure learning of Bayesian networks
  using constraints},'' \emph{{Journal of Machine Learning Research}}, vol.~12,
  no. Mar, pp. 663--689, 2011.

\bibitem{297803:6606182}
M.~Scutari, ``{Learning Bayesian networks with the bnlearn R package},''
  \emph{{arXiv preprint arXiv:0908.3817}}, 2009.

\end{thebibliography}
\end{document}